\documentclass[a4paper,12pt]{article}
\usepackage{epsfig,amsmath,amssymb,amsthm,setspace, amsfonts,cancel,url}
\usepackage[usenames]{color}

\usepackage{booktabs}
\usepackage{tikz}
\usepackage[utf8]{inputenc}
\usepackage{pgfplots} 
\usepackage{pgfgantt}
\usepackage{pdflscape}
\pgfplotsset{compat=newest} 
\pgfplotsset{plot coordinates/math parser=false}
\newlength\fwidth

\usepackage{appendix}
\usepackage{graphicx,subfigure}
\usepackage{multirow}
\usepackage{float}
\usepackage{mathtools}

\newcommand{\condcomment}[2]{\ifthenelse{#1}{#2}{}}

\def \R{\mathbb{R}}
\def\Laplace{\Delta}

\theoremstyle{definition}

\begin{document}
\title{Solving the functional Eigen-Problem using Neural Networks}
\author{Ido Ben-Shaul, Leah Bar and Nir Sochen\\
Department of Applied Mathematics\\
Tel Aviv University, Tel Aviv 69978, Israel\\[0.5in]
}
\maketitle
\numberwithin{equation}{section}

\begin{abstract}
In this work, we explore the ability of NN (Neural Networks) to serve as a tool for finding eigen-pairs of ordinary differential equations. The question we aime to address is whether, given a self-adjoint operator, we can learn what are the eigenfunctions, and their matching eigenvalues. The topic of solving the eigen-problem is widely discussed in Image Processing, as many image processing algorithms can be thought of as such operators. We suggest an alternative to numeric methods of finding eigenpairs, which may potentially be more robust and have the ability to solve more complex problems. In this work, we focus on simple problems for which the analytical solution is known. This way, we are able to make initial steps in discovering the capabilities and shortcomings of DNN (Deep Neural Networks) in the given setting.
\end{abstract}

\section{Introduction}
Eigen-functions and eigen-values of the Laplacian (and other operators) are important in various applications ranging from image processing, computer vision and shape analysis. It is also of major important in various engineering applications where resonances are of crucial for design and safety. We present in this paper a new numerical method for the computation of these eigenfunctions and eigenvalues. The resulting eigenfunctions are {\em smooth} functions defined in a parametric way. This is in contrast to finite element or finite difference method in which the eigenfunction is defined on a grid or is piecewise linear or polynomial with a limited smoothness. 

This work is an extension of the work \cite{BarSochen} where unsupervised Deep Neural Networks (DNNs) are applied to solve differential equations and in particular the Electrical Impedence Tomography (EIT) problem.  We are extending that work to the case of eigenfunction differential equation with unknown eigenfunctions and eigenvalues.  

We demonstarte our new method on the very simple case of the one-dimensional Laplacian on an interval where we have the analytical ground truth solution to compare with.   

\section{Preliminaries}
Let ${\cal{H}}$ be a Hilbert space where the inner product for $u,v\in {\cal{H}}$ is $(u,v)$. Let $A\in O({\cal{H}})$ be an operator. Let $A^*$ be the adjoined operator defined by $(A^*u,v)=(u,Av)\ \  \forall u,v\in {\cal{H}}$. Then $A$ is said to be self-adjoint, if $A=A^*$.
We start off with a short Lemma on self-adjoint operators \cite{JBConway}.
Let ${\cal{H}}$ be a Hilbert space. Let $A\in O({\cal{H}})$ be a self-adjoint operator. Then all eigenvalues of A are real.

In this article, we limit ourselves to self-adjoint operators, and therefore searching for eigenpairs: $(\lambda, u)$ s.t. $\lambda \in \R.$ We begin by looking at a solution proposed in \cite{haitfraenkel2019revealing}, where an algorithm is given based on numerical methods, and generalizations of the Power Method for blackbox operators ??? (instead of the simple, linear case). This method requires operator $ T: {\cal{L}}_2(\R^n) \rightarrow {\cal{L}}_2(\R^n)$.  We decided to focus on $ T: {\cal{L}}_2(\R) \rightarrow {\cal{L}}_2(\R)$. In this way, the methods are comparable. Notice, there are no restrictions on the functional for using our method: $ T: {\cal{L}}_2(\R^n) \rightarrow {\cal{L}}_2(\R^k)$. We have used the well known Laplace Equation: $Tu=\Delta u$.


\section{Method} 
We introduce several tactics at pursuing the solution to the Eigen-Problem. Our approach follows \cite{BarSochen}, using the initial Loss Function introduced there:
\begin{equation}
\mathcal{F}(u) = \alpha\|\mathcal{T}u\|_{2}^2+\mu\|\mathcal{T}u\|_{\infty}+\delta\|u-u_0\|_{1,\partial\Omega}+\mathcal{R}^F(u),
\label{eq:Fu}
\end{equation}
where in our case: $Tu=\Delta u + \lambda u$. For later use we define 
\begin{equation}
\mathcal{F}_1(u) = \alpha\|\mathcal{T}u\|_{2}^2+\mu\|\mathcal{T}u\|_{\infty}+\delta\|u-u_0\|_{1,\partial\Omega}\ .
\label{eq:F1u}
\end{equation}
Of course, in the general case, the eigenvalue is not given. Yet, as a first challenge, we permit ourselves the knowledge of a specific eigenvalue - $\lambda$, and ask if we can solve the eigen-problem (there is an infinite amount of eigenpairs for the laplace equation).

Due to the fact that an eigenfunction is invariant to dilations - i.e. for $(u,\lambda)$ an eigen-pair for functional $T$, the pair $(\alpha u, \lambda) $ is also an eigen-pair of $T$, we wish to fix the \textit{energy} of the eigenfunction, and in this way have a single solution for every eigenvalue. We introduce another term to the loss function:
\begin{equation}
E_{penalty}= \beta|\|u\|_{2}^2-c|,
\label{eq:Epenalty}
\end{equation}
\vspace{.12 in}
where $c$ is the desired Energy, and $\|.\|_{2}$ is defined as the the Monte-Carlo Integral approximation of the $L_{2}$ norm:
\begin{equation}
\|u\|_{2}= \frac{b-a}{N} \sum_{x=x_{1}}^{x_{N}}(u(x))^2 \sim \int_{a}^{b} (u(x))^2 dx
\label{eq:Energy}
\end{equation}
and in general, we will approximate:
\begin{equation}
\langle f,g\rangle_{2}= \frac{b-a}{N} \sum_{x=x_{1}}^{x_{N}}\bar{f}(x)g(x) \sim \int_{a}^{b} \bar{f}(x)g(x) dx
\label{eq:innerP}
\end{equation}

\subsection{An eigenpair with known eigenvalue}

This sectiont aims to answer the following question: given an eigenvalue $\lambda$, can we approximate the matching eigenfunction using a NN? We give three different eigenpairs, with different boundary conditions and show the learned function as opposed to the GT (Ground Truth) solution :

\begin{center}

\begin{figure}[H]
\includegraphics[width=0.30\textwidth]{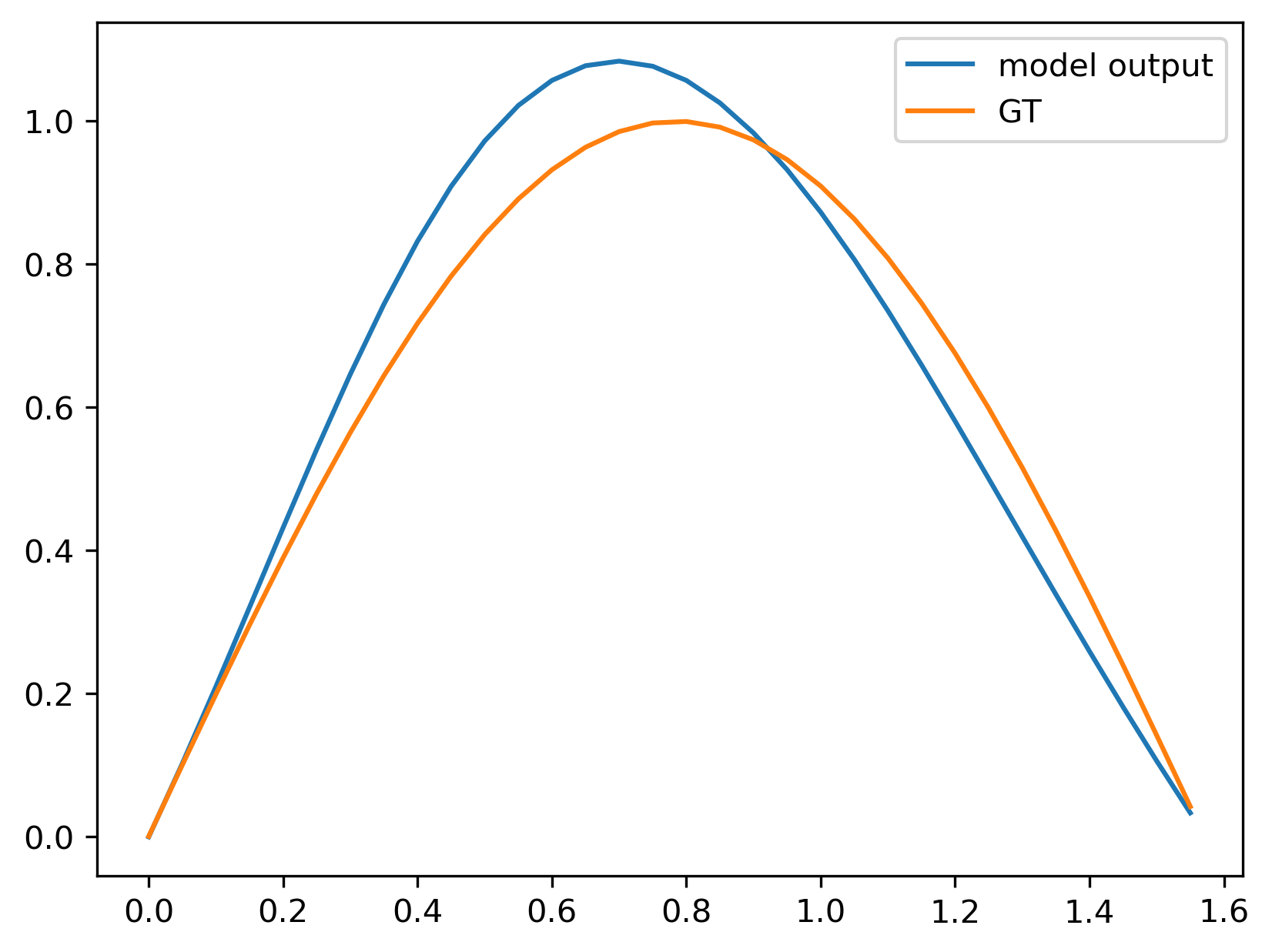}
\hfill
\includegraphics[width=0.30\textwidth]{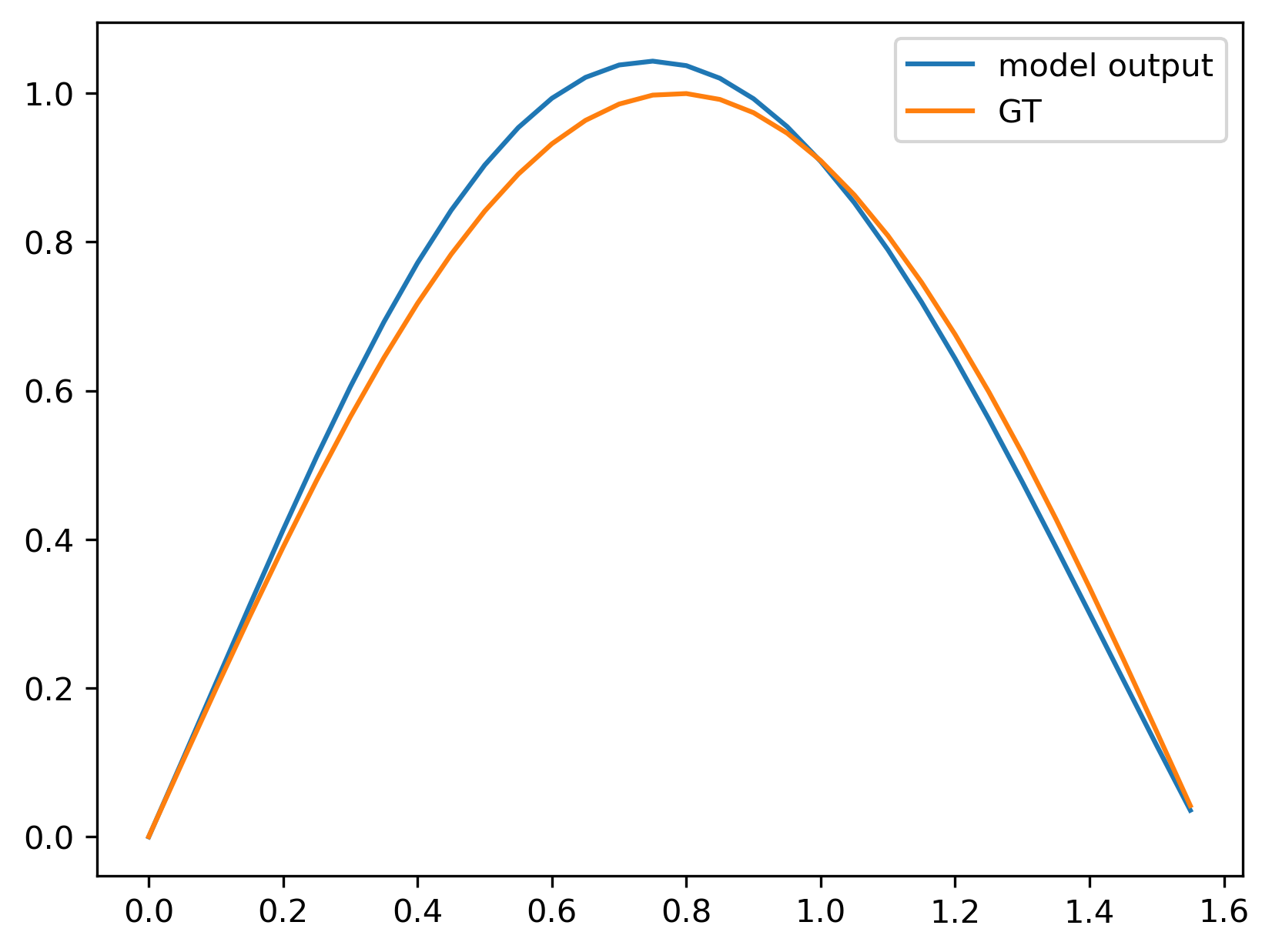}
\hfill
\includegraphics[width=0.30\textwidth]{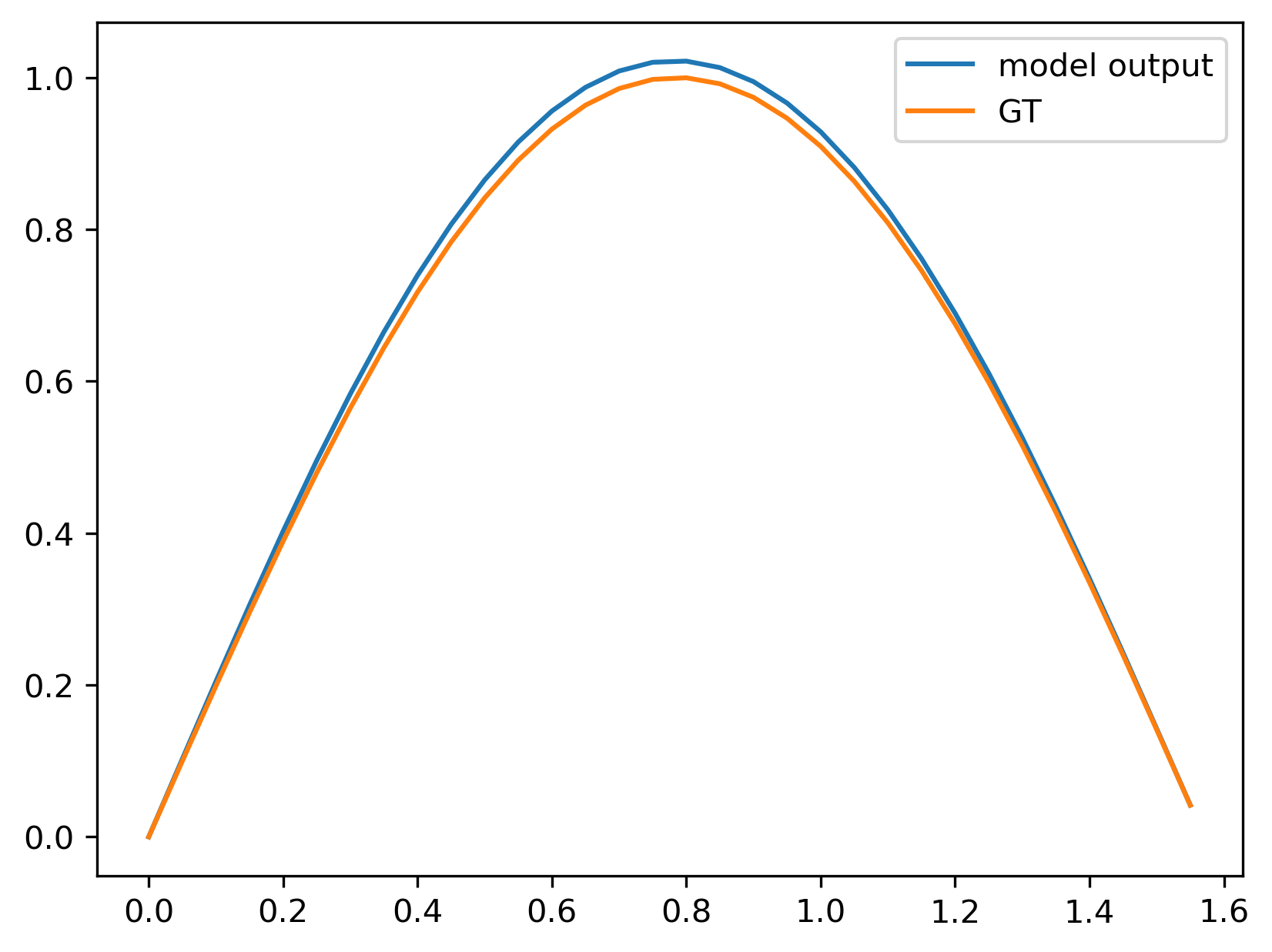}
\caption{$u'' + 4u = 0, u(0) =0, u(\pi/2) = 0,$ Solution: $u=sin(2x)$, $Energy = 1$. The graphs show the learned function and the GT function at epochs 100, 500, 1000 - from left to right. }
\end{figure}


\begin{figure}[H]
\includegraphics[width=0.30\textwidth]{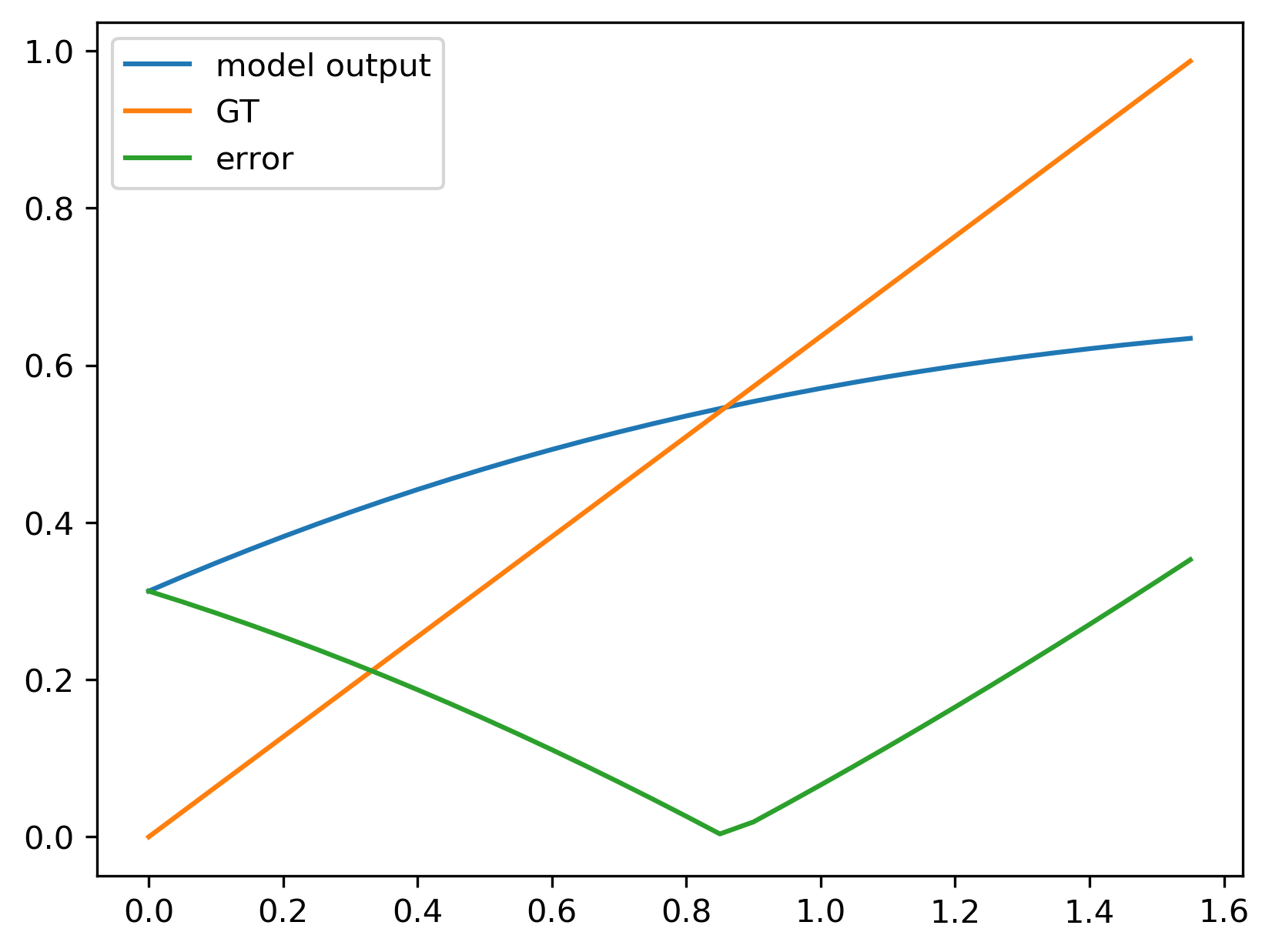}
\hfill
\includegraphics[width=0.30\textwidth]{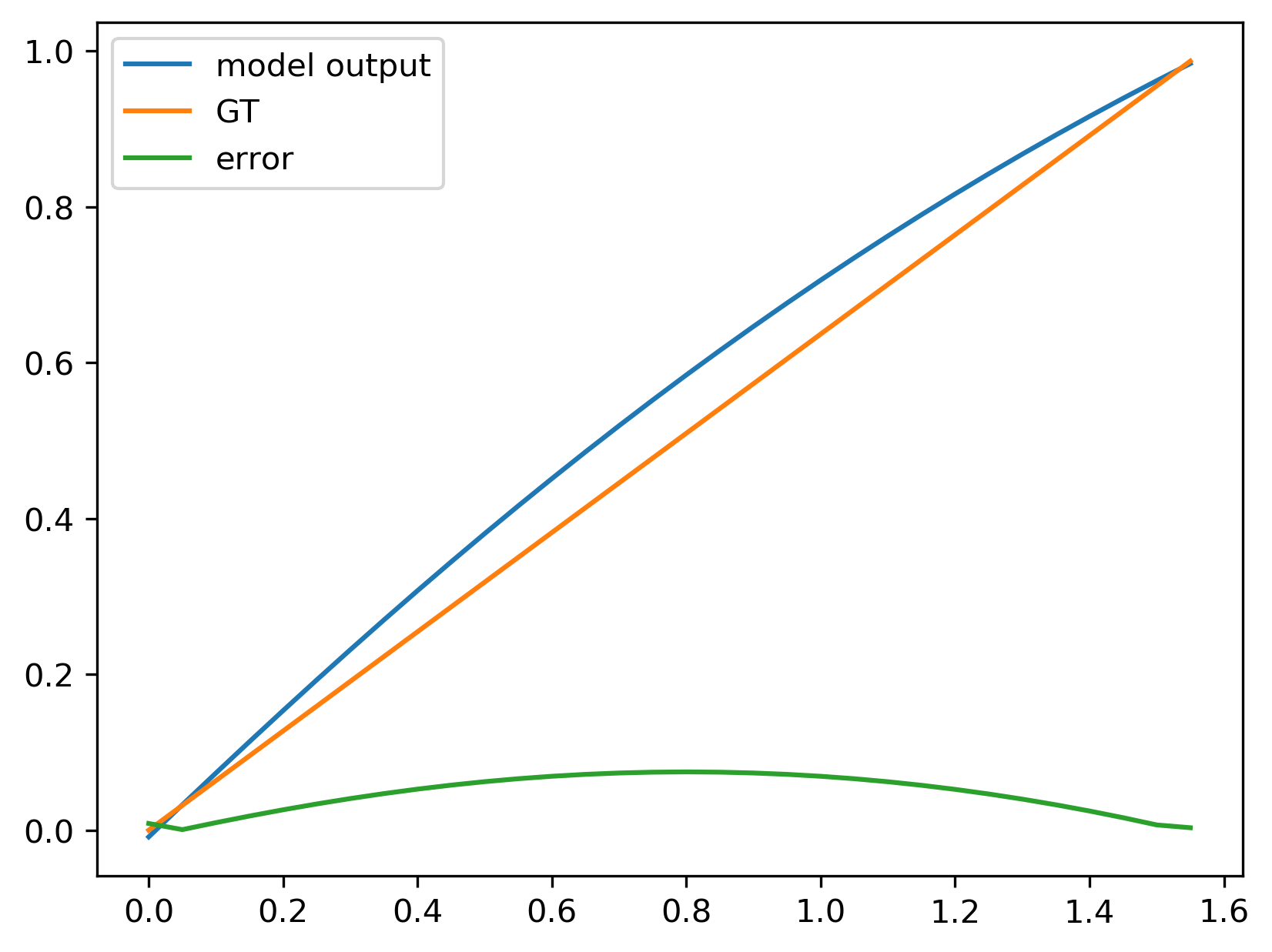}
\hfill
\includegraphics[width=0.30\textwidth]{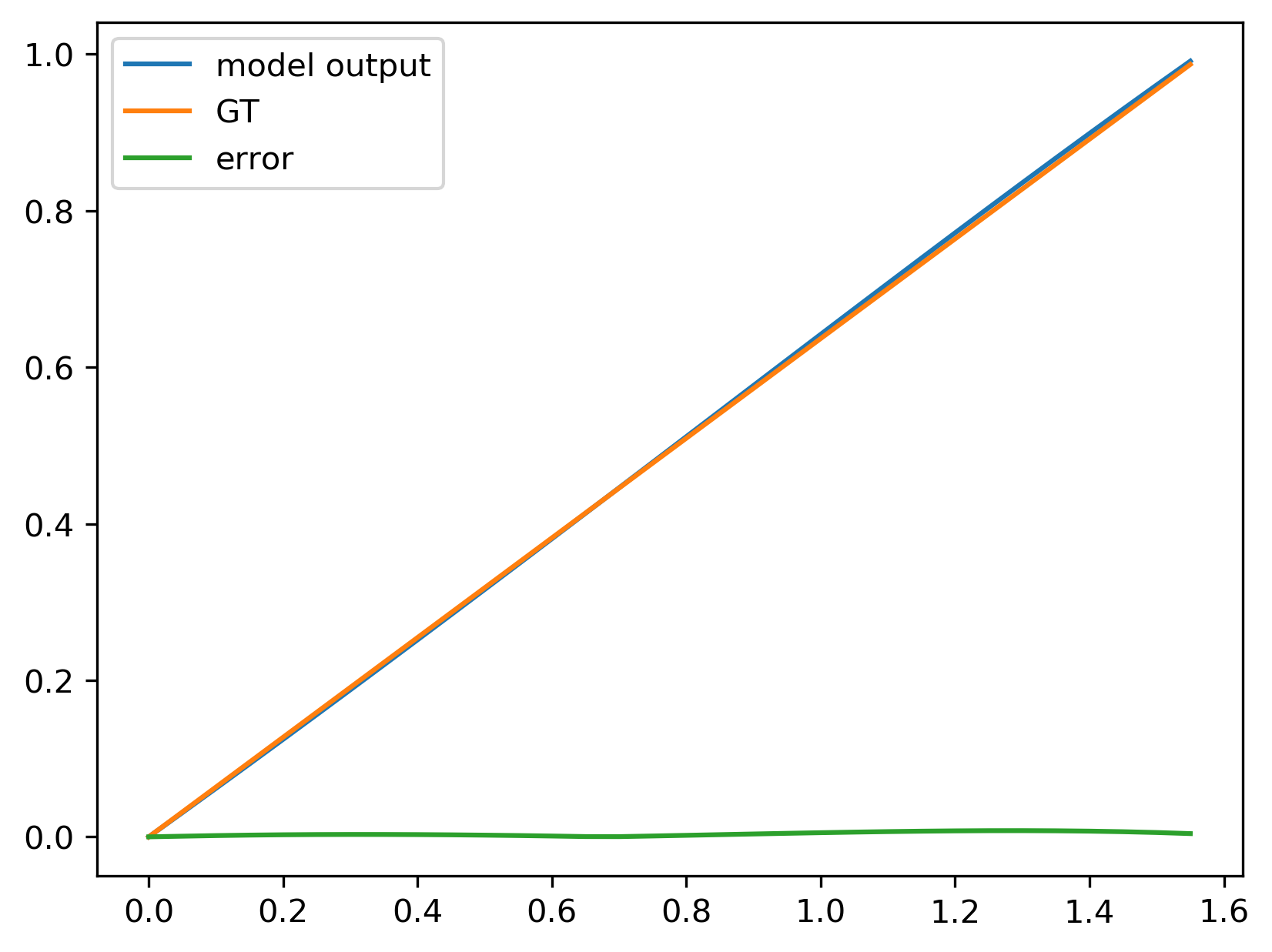}
\caption{$u''= 0, u(0) =0, u(\pi/2) = 1,$ Solution: $u = \frac{2x}{\pi}, Energy = \pi/4$. The graphs show the learned function, the GT function, and the Error at epochs 1, 50, 100 - from left to right.}
\end{figure}


\begin{figure}[H]
\includegraphics[width=0.30\textwidth]{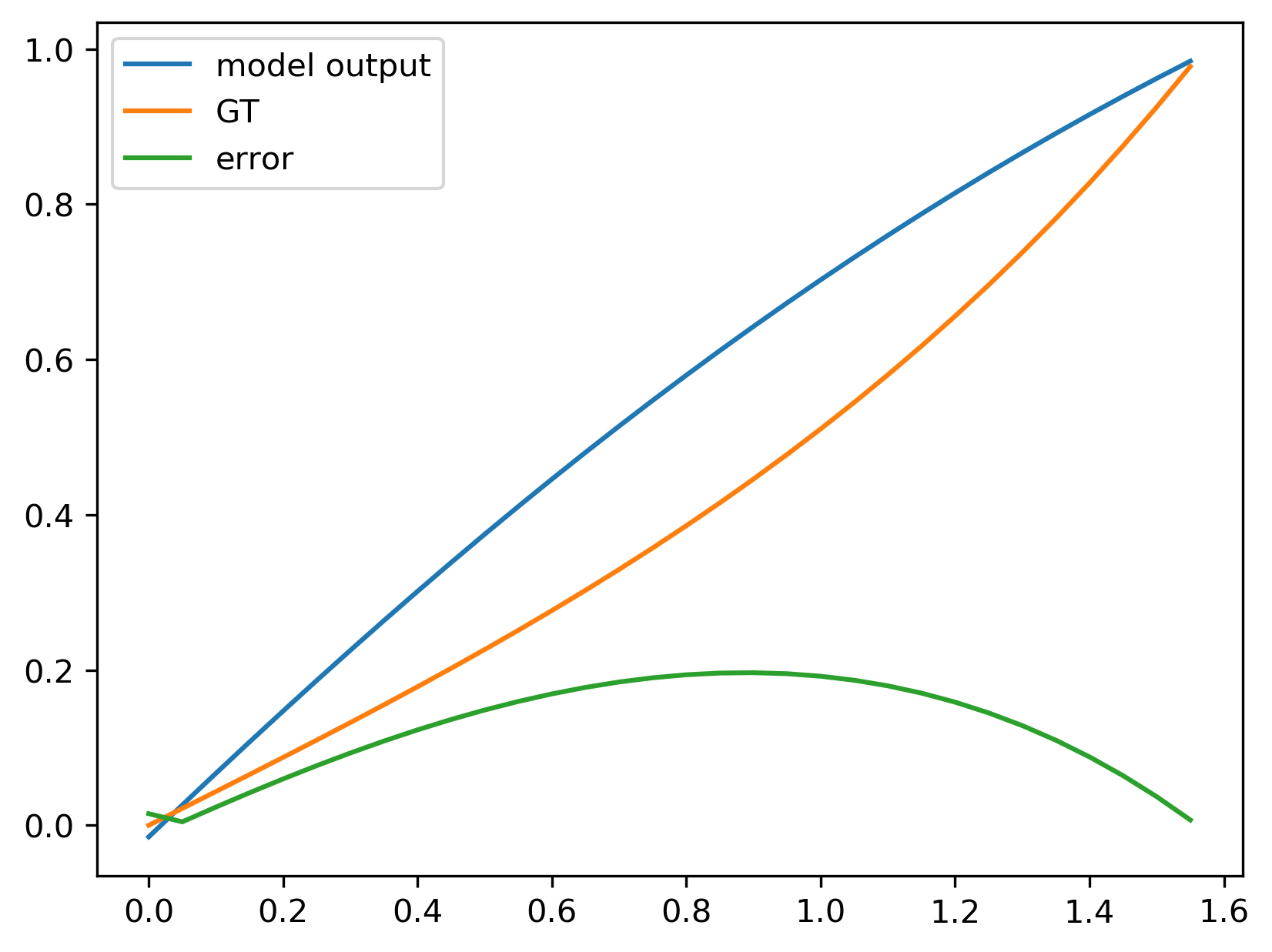}
\hfill
\includegraphics[width=0.30\textwidth]{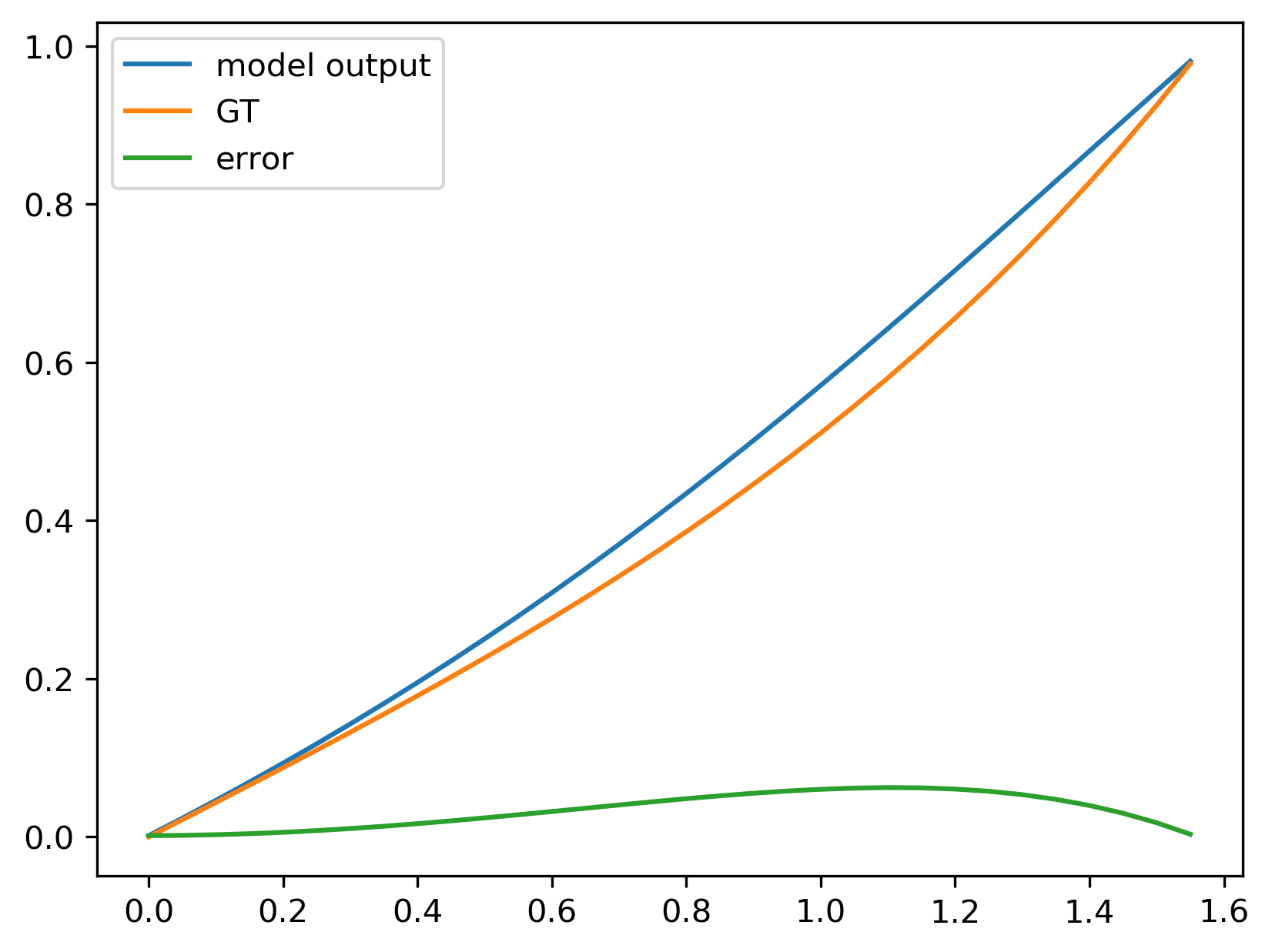}
\hfill
\includegraphics[width=0.30\textwidth]{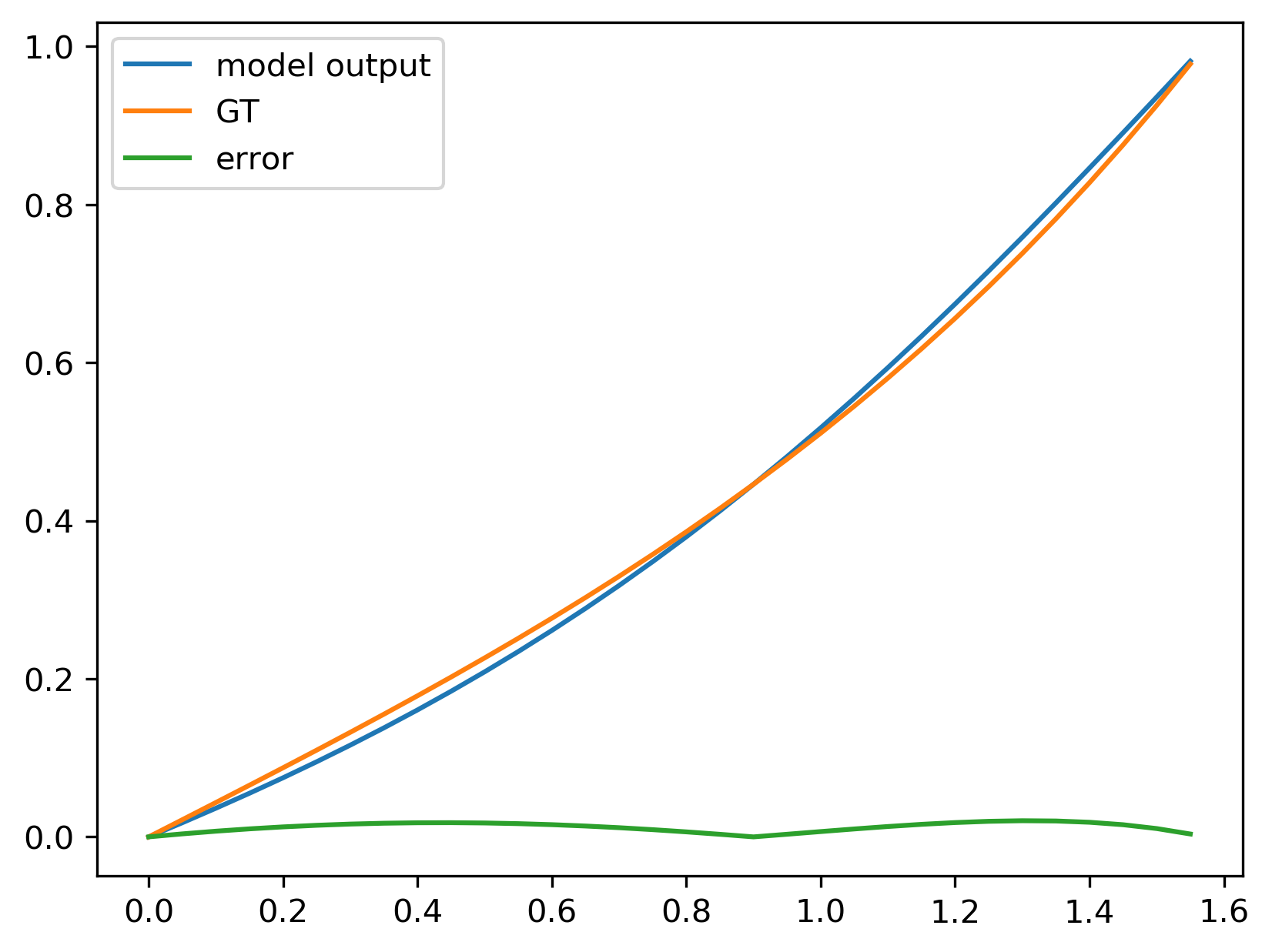}
\caption{$u''- u= 0, u(0) =0, u({\pi}/2) = 1,$ Solution: $u = \frac{{e^{\frac{\pi}{2}}-x}(e^{2x}-1)}{(e^{\pi}-1)}, Energy: tanh(\pi/4)$. The graphs show the learned function, the GT function, and the Error at epochs 10, 20, 100 - from left to right.}
\end{figure}
\end{center}

In general, for the rest of the article, we keep our the energy conditions to be $1$ for all eigenfunctions.

\subsection{One eigenpair}

Our next challenge was incorporating the eigenvalue search into the neural network, to be learned. For this purpose, we used the well known Rayleigh Quotient \cite{GLMiller} \cite{Feld_2019}. 

\begin{equation}
R(u) := \frac{-<Tu, u>}{<u, u>}
\end{equation}

It can be shown that for a given eigenfunction $u, R(u)$ is the matching eigenvalue, and furthermore, functions nearing an eigenfunction will have a Rayleigh Quotient close to the matching eigenvalue. In other words, the Rayliegh Quotient answers the question: Given $x$, what scalar $\alpha$
"acts most like an eigenvalue" for $x$ in the sense of minimizing $|\|Tx + \alpha x\|_{2}$ ? An example proof is given in \cite{GLMiller}. Using this rationale, the sought eigenfunction $u$, now holds:

\begin{equation}
Tu = \Laplace u + R(u)u
\end{equation}

Given the fact that there are an infinite number of eigenvalues, we add a penalty to the size of the Rayleigh Quotient, $\gamma|\R(u)\|_{2}^2$. 
Our loss, then, becomes of the following form:

\begin{equation}
\mathcal{F}(u) = \mathcal{F}_1(u)+\beta|\|u\|_{2}^2-c|+\mathcal{R}^F(u) + \gamma|\|R(u)\|_{2}^2,
\end{equation}

Given this setting, we are still looking at the problem: $u''=0,   u(0)=0, u(\pi)=0, E(u)=1$, and searching for eigenpairs. The analytic solution is $u=\frac{2}{\pi}sin(kx), \lambda=k^2, k \in \mathbb{N}$. Using the loss defined above, the NN is able to focus on the smallest eigenvalue, and learn its matching eigenfunction:

\begin{center}
\begin{figure}[H]
\centering
\includegraphics[width=0.30\textwidth]{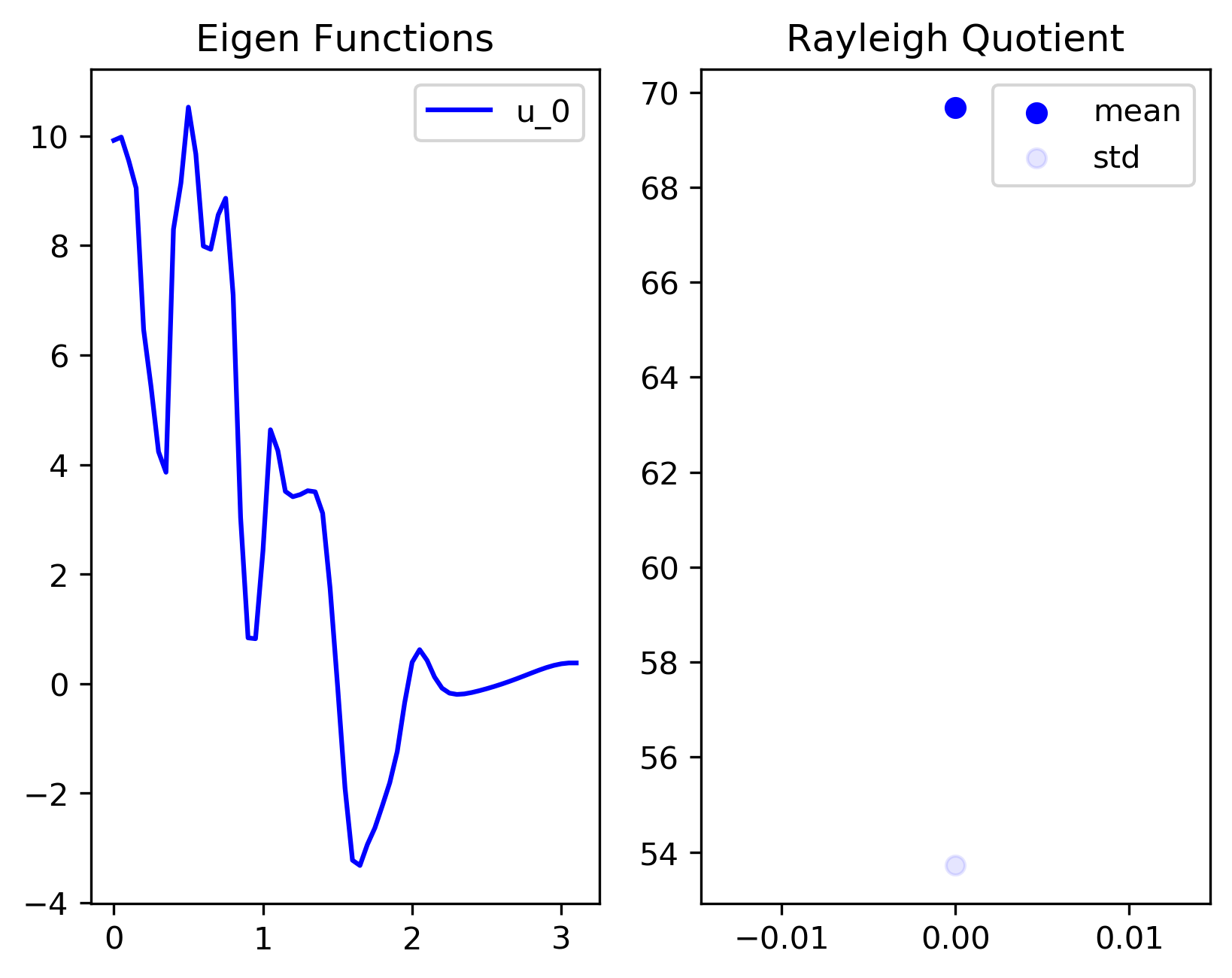}
\hfill
\includegraphics[width=0.30\textwidth]{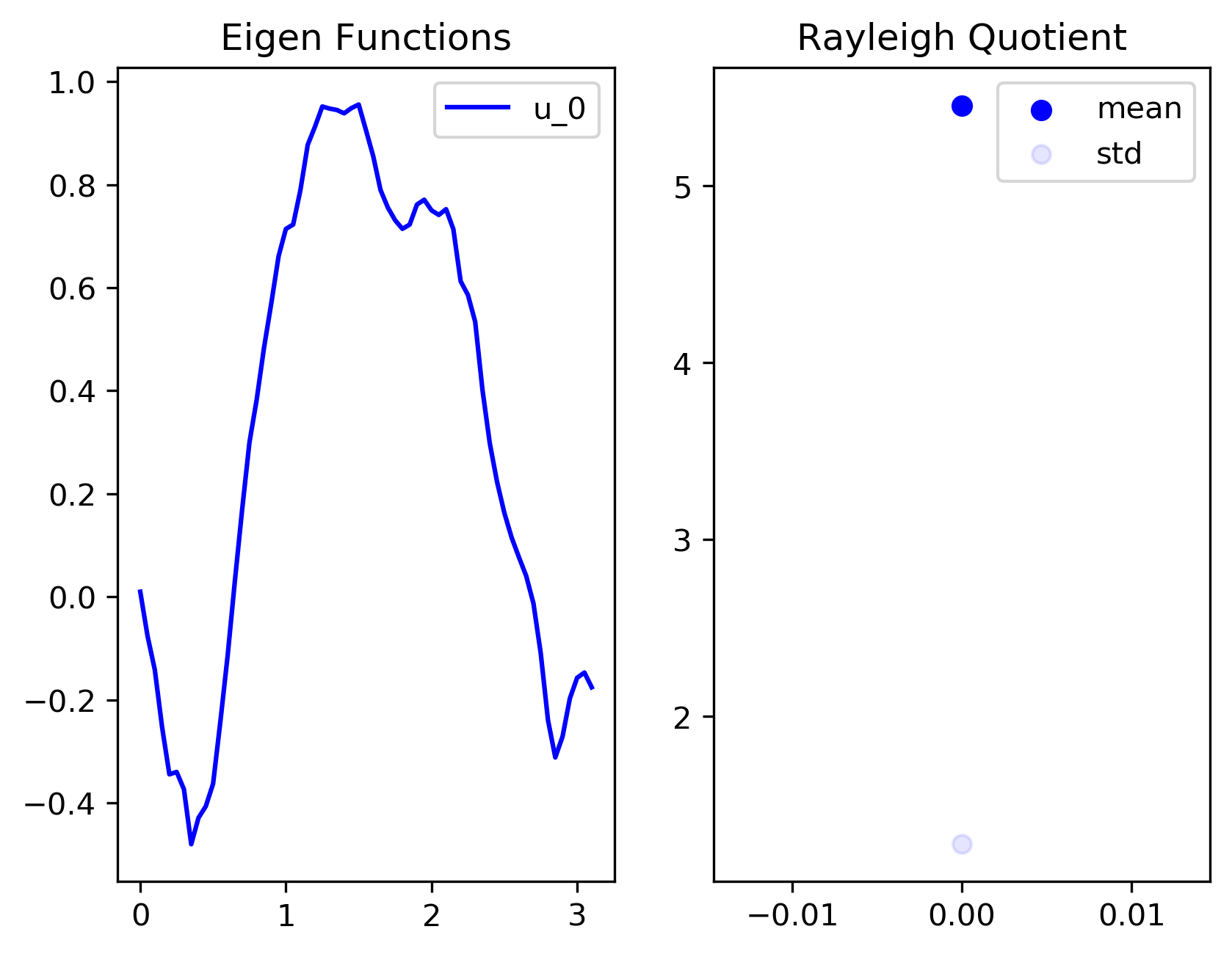}
\hfill
\includegraphics[width=0.30\textwidth]{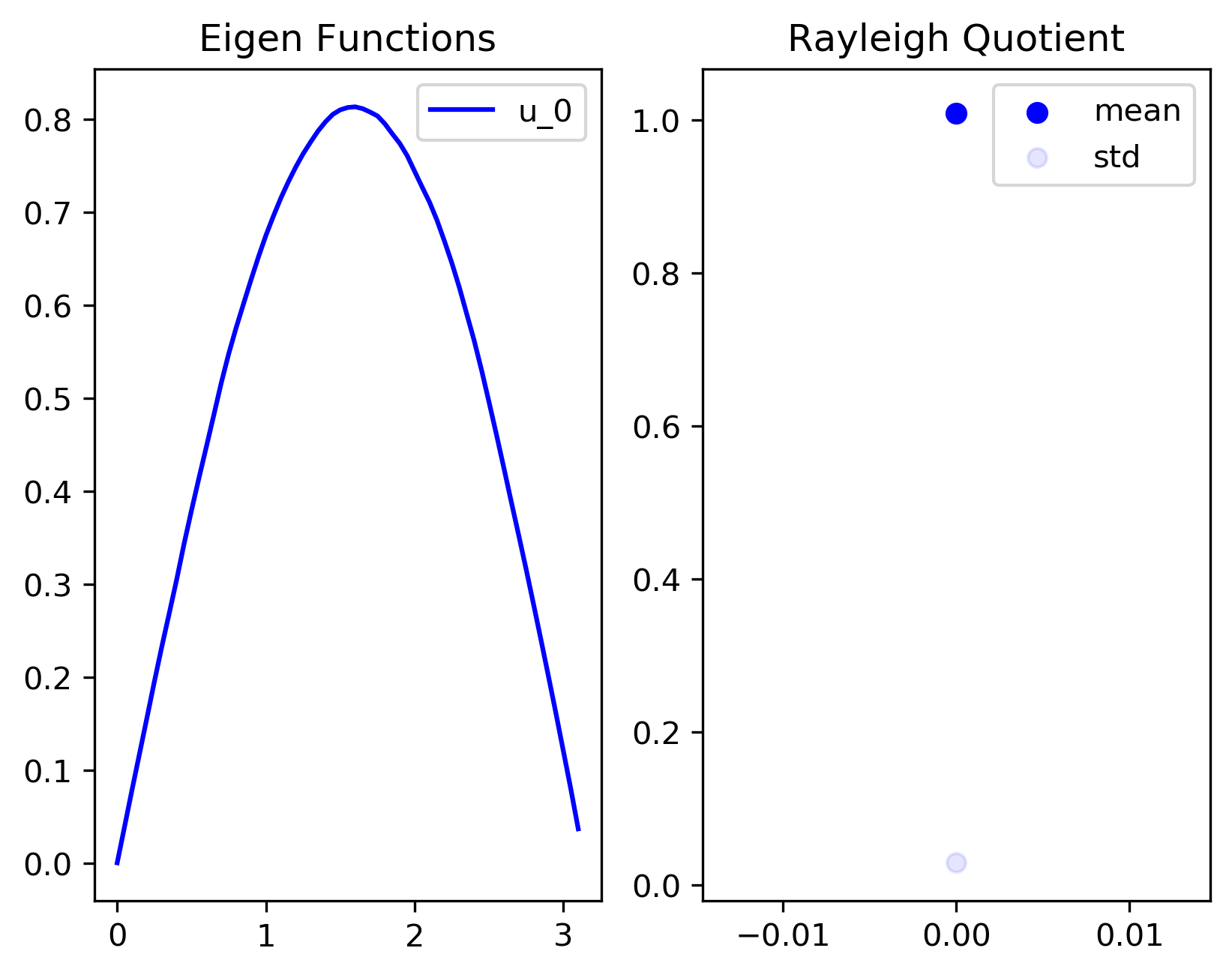}
\caption{Learning of the smallest eigenpair: $(1, \frac{2}{\pi}sin(x))$. Three sampled epochs are shown at 1, 50, 1000 - from left to right. In the right graph, we can see the mean and std($\sigma$)  of the matching Rayleigh Quotient(Per Batch). It is clear that the Rayleigh Quotient indeed converges to the wanted value, and its $ \sigma  \rightarrow 0$ as the epochs progress.}
\end{figure}
\end{center}

\subsection{Several eigenpairs}

Our last task was to generalize the now proven method of finding one eigen-pair and find many {\bf different} eigen-pairs. In order to accomplish this task, we introduce an orthogonality term, similar to the approach done in \cite{haitfraenkel2019revealing}. Our network now has 1 input, and $k$ outputs- as the number of desired eigenpairs. The Loss will be defined as follows:

\begin{equation}
\mathcal{F}_s(u) = \sum_{i=1}^{m}(\mathcal{F}_1(u_i)+\beta|\|u_i\|_{2}^2-c|+\mathcal{R}^F(u_i) + \gamma_{i}|\|R(u_i)\|_{2}^2) + \nu\sum_{i<j}^{}\langle u_{i},u_{j}\rangle
\label{eq:Fsu}
\end{equation}

We keep the Rayleigh penalty term, to receive the $k$ smallest eigenpairs. Our $ \gamma_{i}$ need not be the same. We give the following results, for the problem given above - Laplace Equation with Dirichlet Boundary Conditions. 

\begin{center}
\begin{figure}[H]
\includegraphics[width=0.32\textwidth]{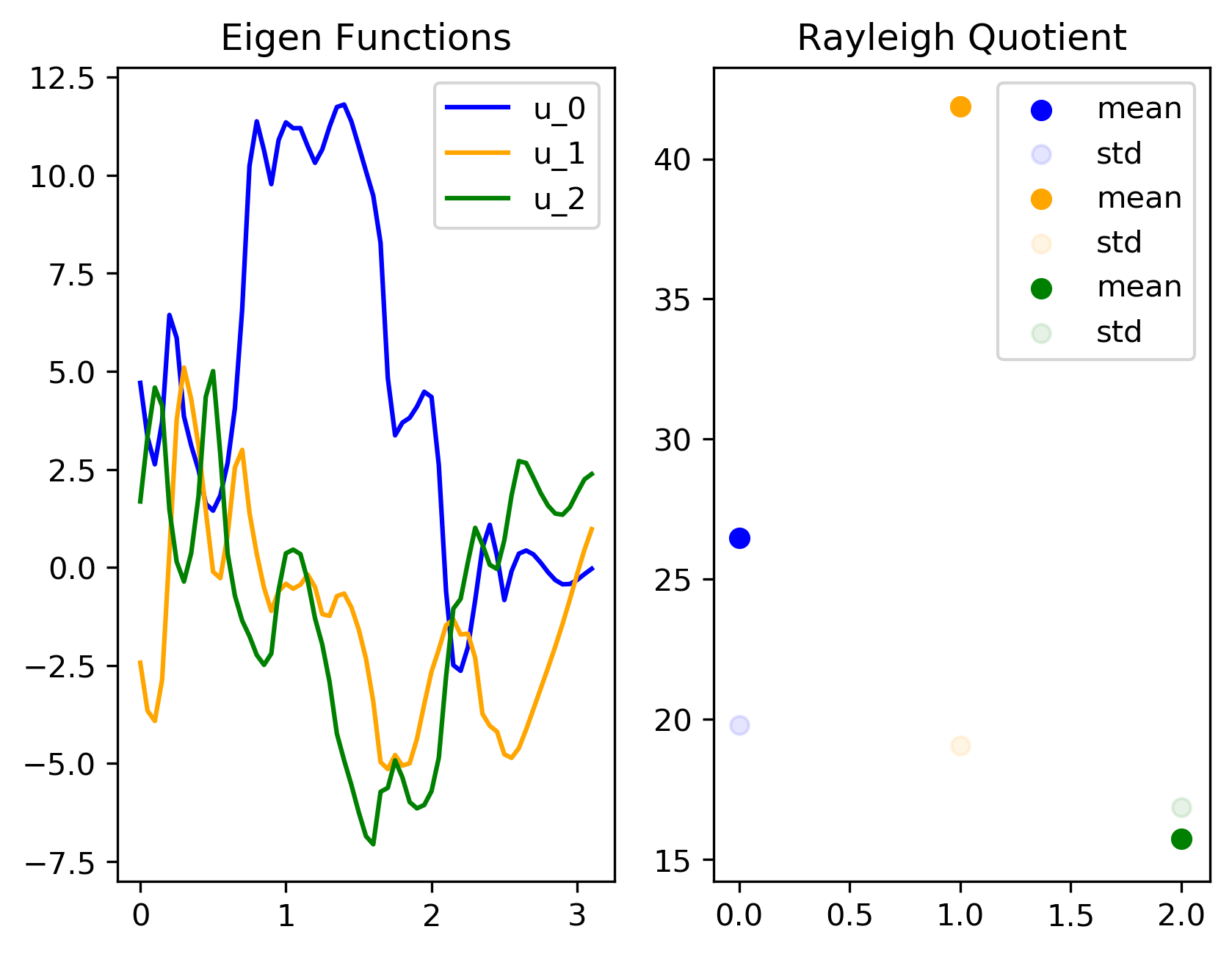}
\hfill
\includegraphics[width=0.32\textwidth]{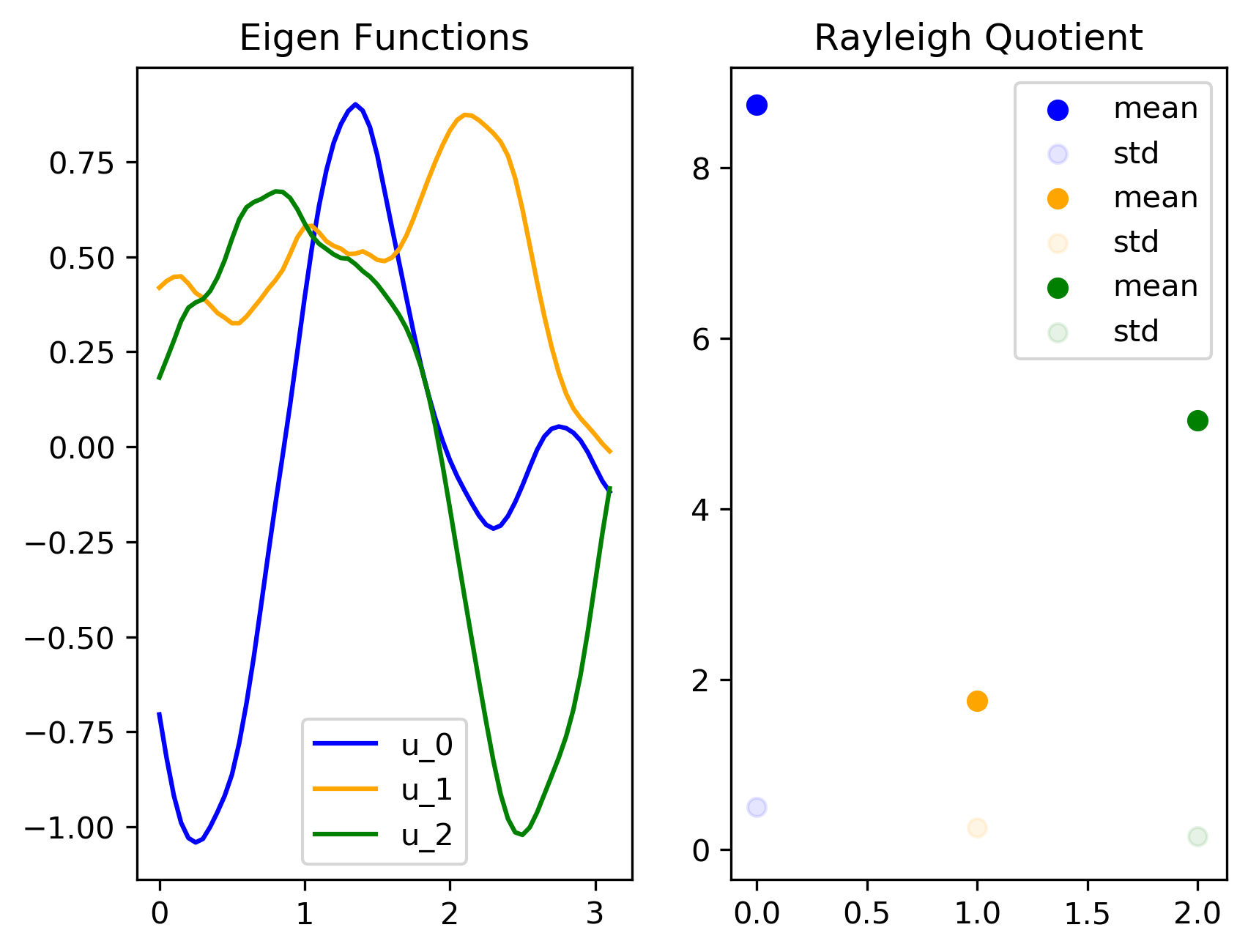}
\hfill
\includegraphics[width=0.32\textwidth]{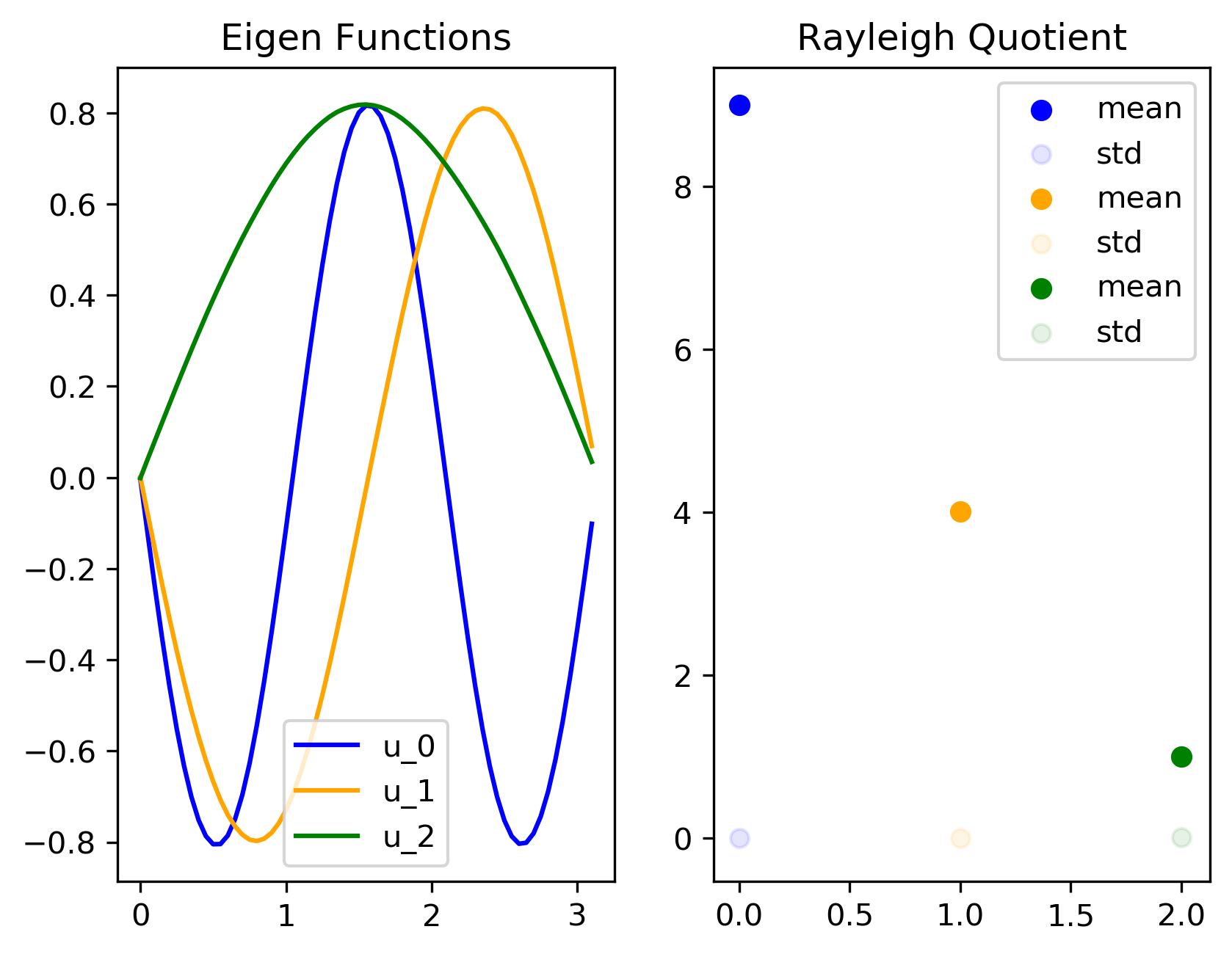}
\caption{Learning 3 different eigenpairs.  Three sampled epochs are shown at 1, 50, 5000 - from left to right. We can see that the learned eigenvalues match our analytic solution- namely $[1, 4, 9]$, and that their $\sigma\rightarrow0$. We can also see that the learned eigenfunctions are indeed the analytic solution, up to $sign$(as expected).}
\end{figure}
\end{center}

We also experimented with different Rayleigh Quotient penalizing. One approach is taking a factor $\frac{1}{i} $ for $i={1...m}$ . With this approach, we were able to learn more eigenfunctions. 

\begin{center}
\begin{figure}[H]
\centering
\includegraphics[width=0.25\textwidth]{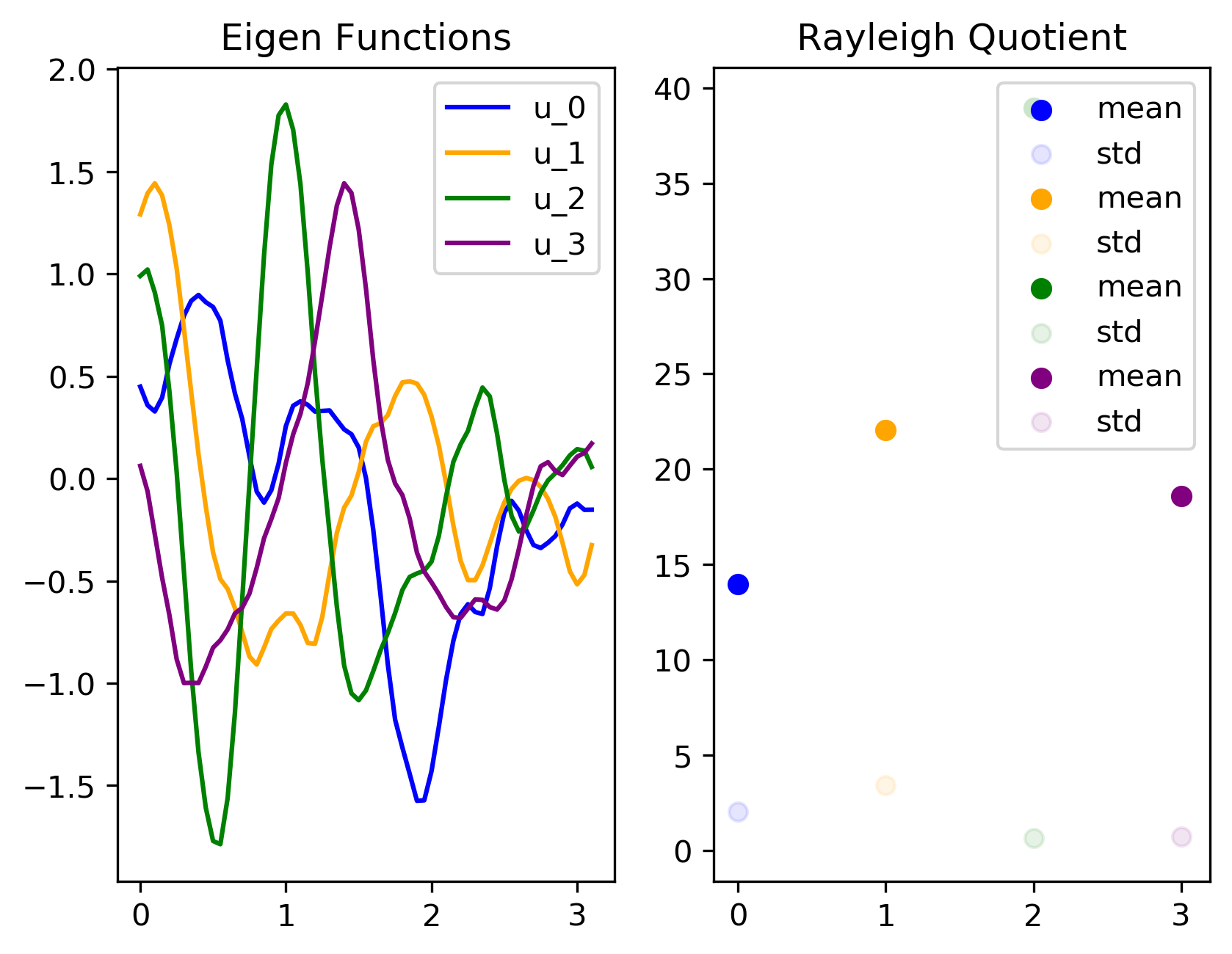}
\hfill
\includegraphics[width=0.25\textwidth]{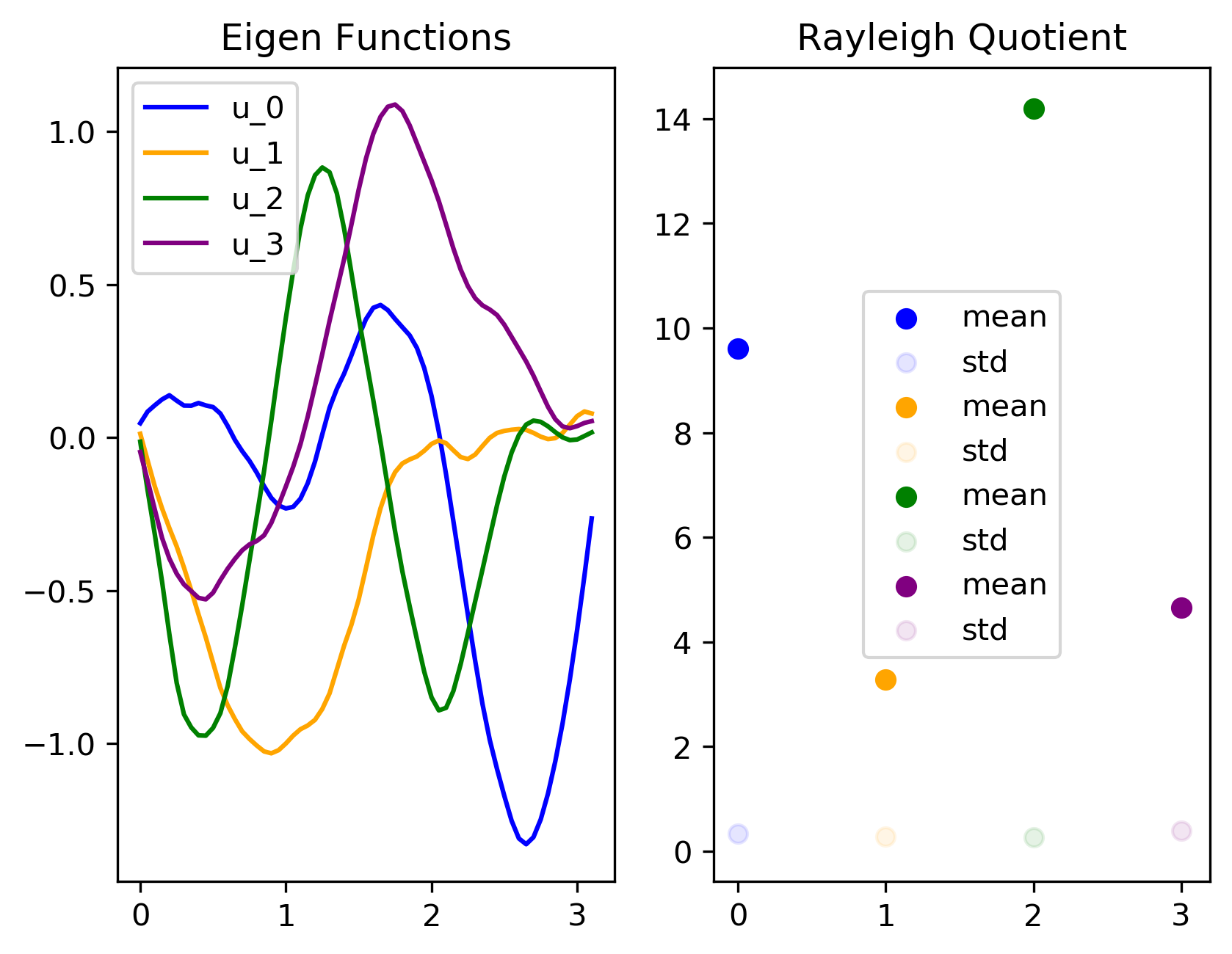}
\hfill
\includegraphics[width=0.25\textwidth]{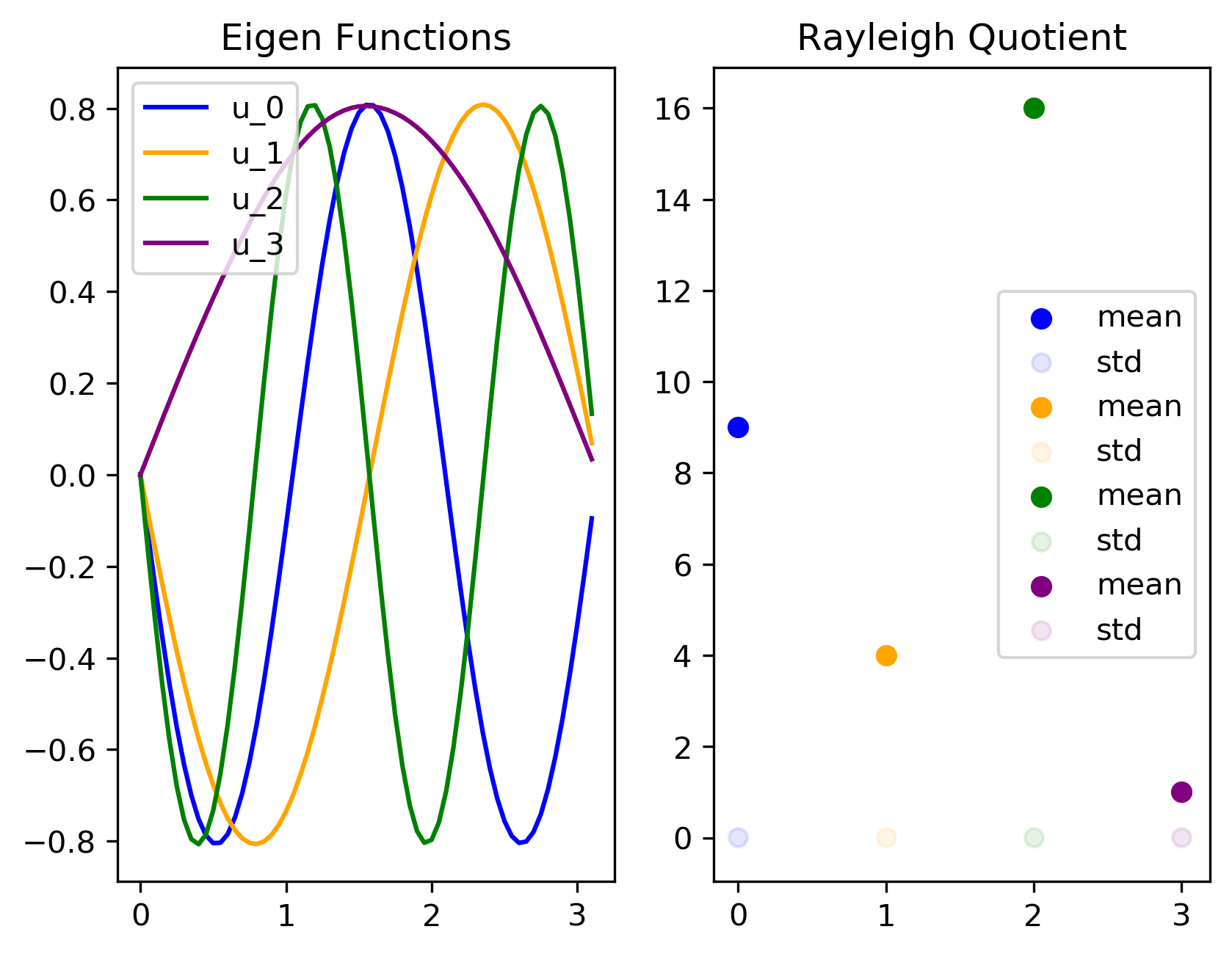}
\hfill
\includegraphics[width=0.3\textwidth]{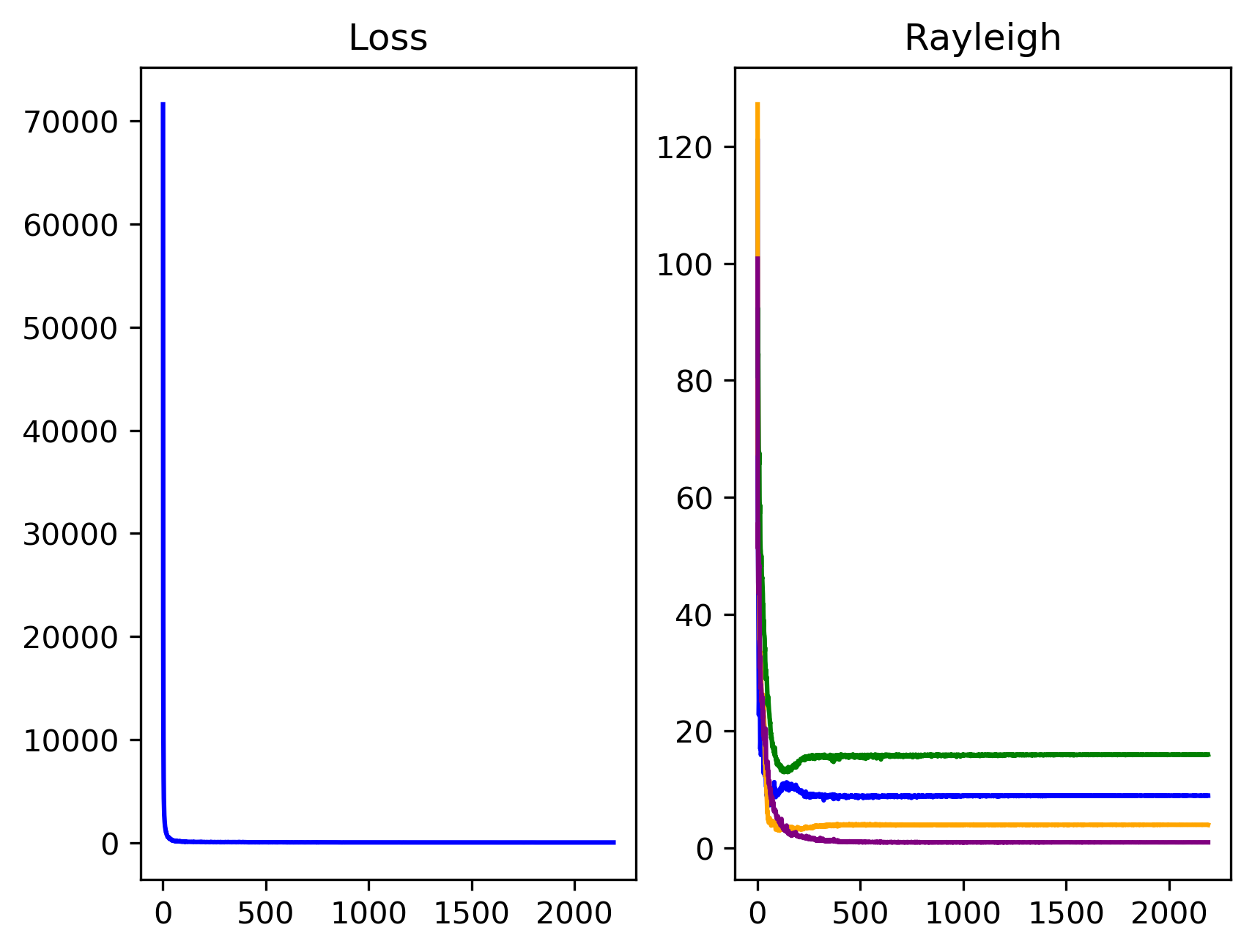}
\caption{Learning 4 different eigenpairs using factor weighted Rayleigh Quotient Penalty Loss. Three sampled epochs are shown at 50, 200, 3000 - from left to right. The bottom graph shows the Training Loss, and the Rayleigh Quotient - mean and std($\sigma$) - per epoch. It is clear that the Loss quickly stabilizes, as well as each of the different eigenvalues- matching different eigenfunctions.}
\end{figure}
\end{center}

\begin{center}
\begin{figure}[H]
\centering
\includegraphics[width=0.25\textwidth]{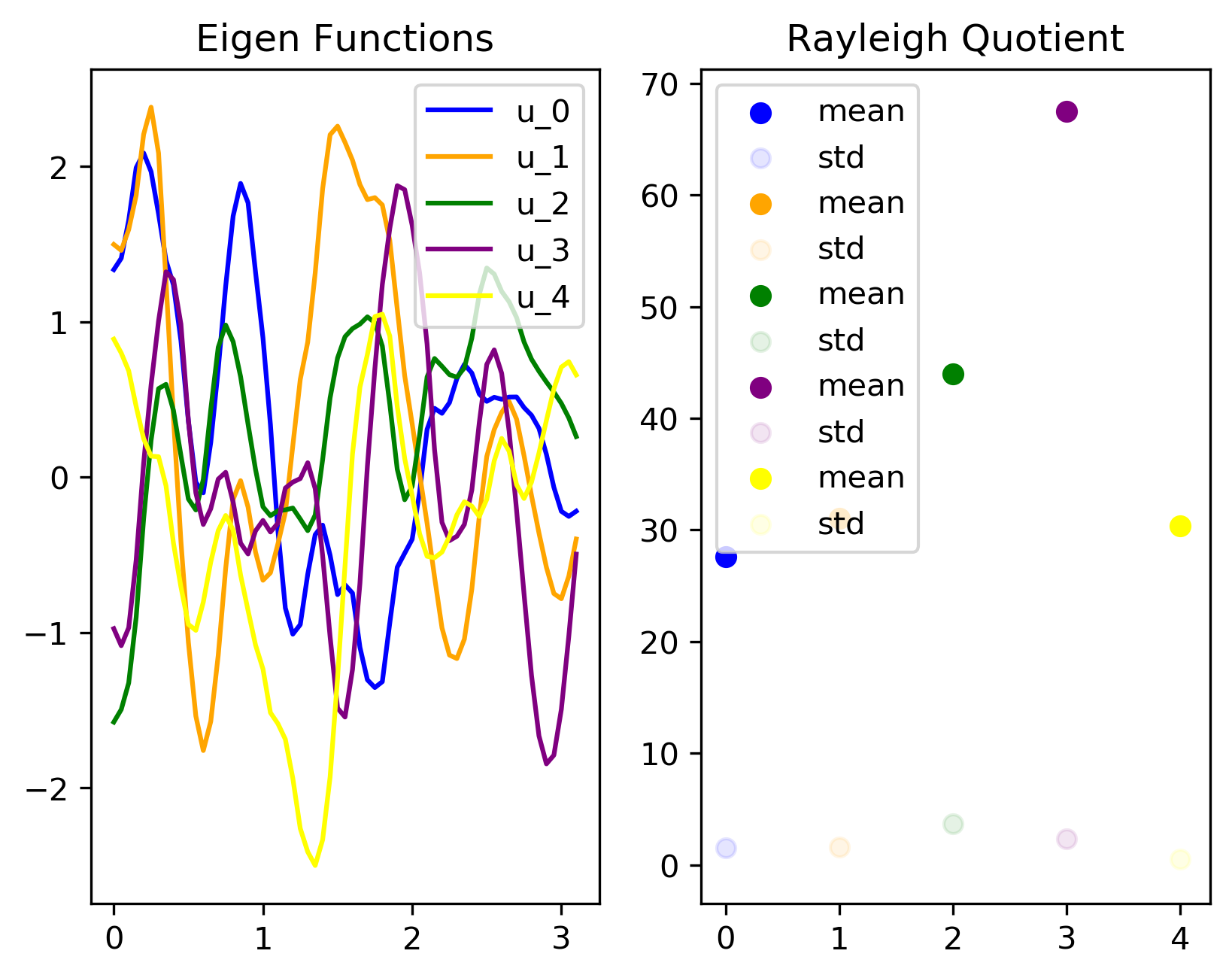}
\hfill
\includegraphics[width=0.25\textwidth]{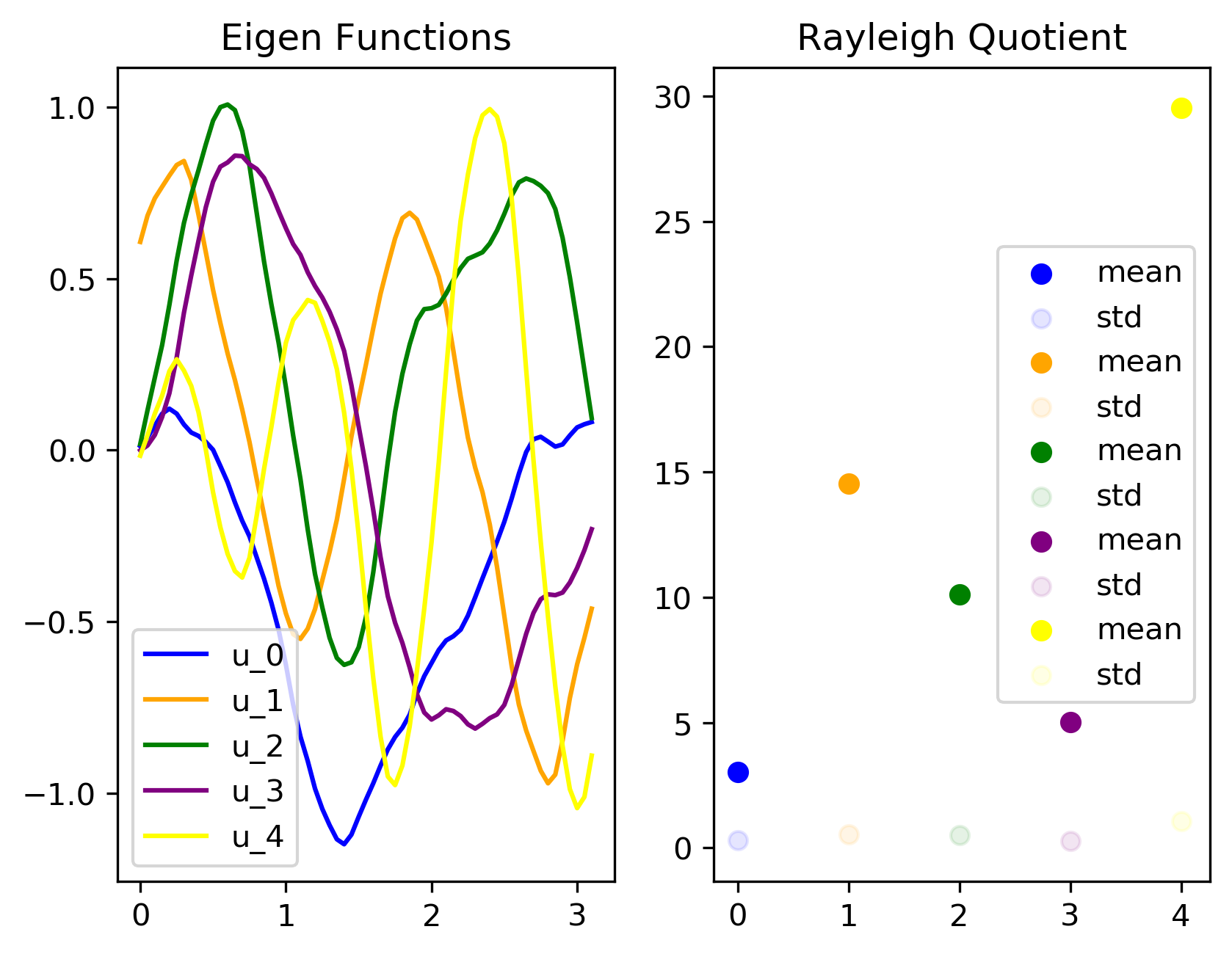}
\hfill
\includegraphics[width=0.25\textwidth]{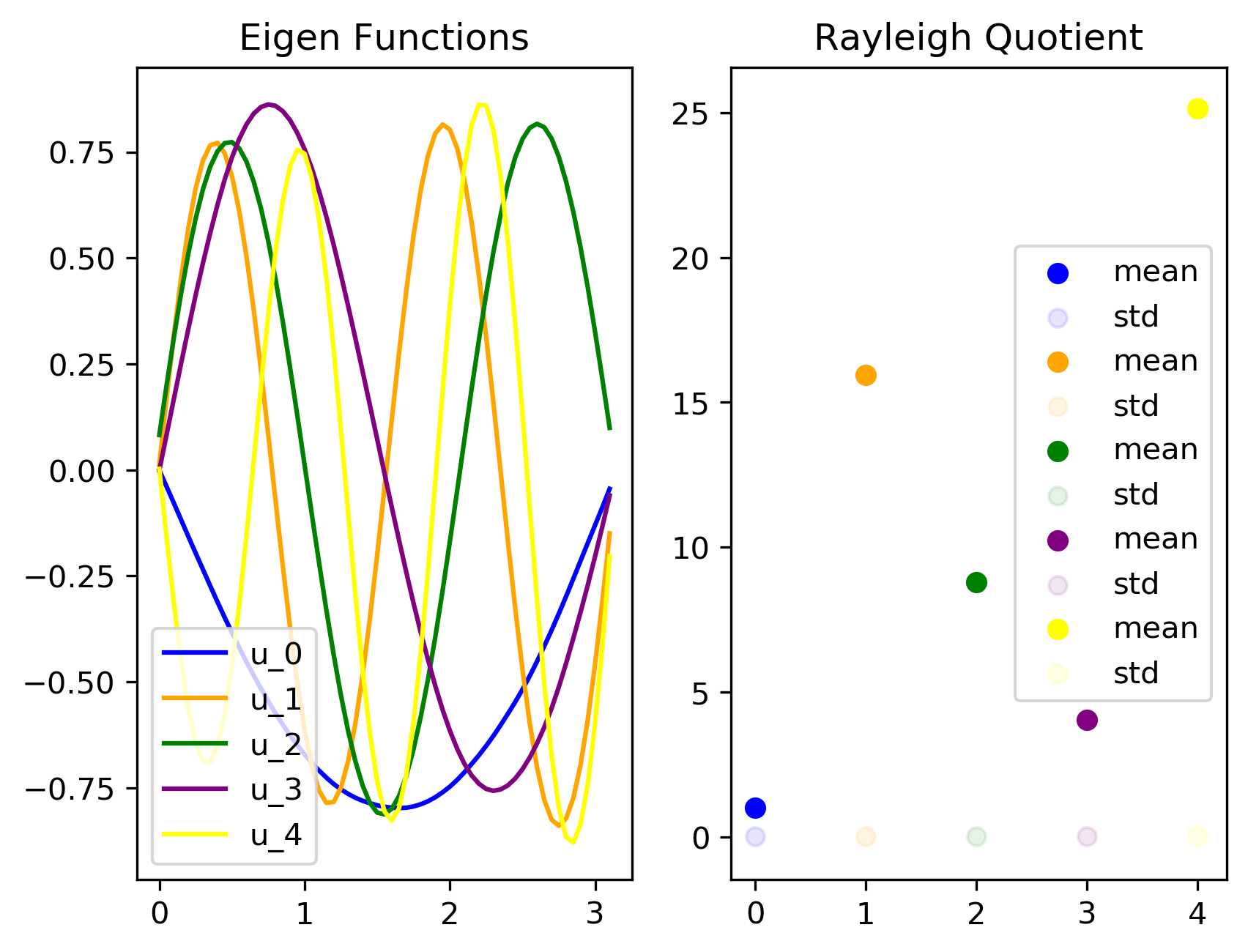}
\hfill
\includegraphics[width=0.3\textwidth]{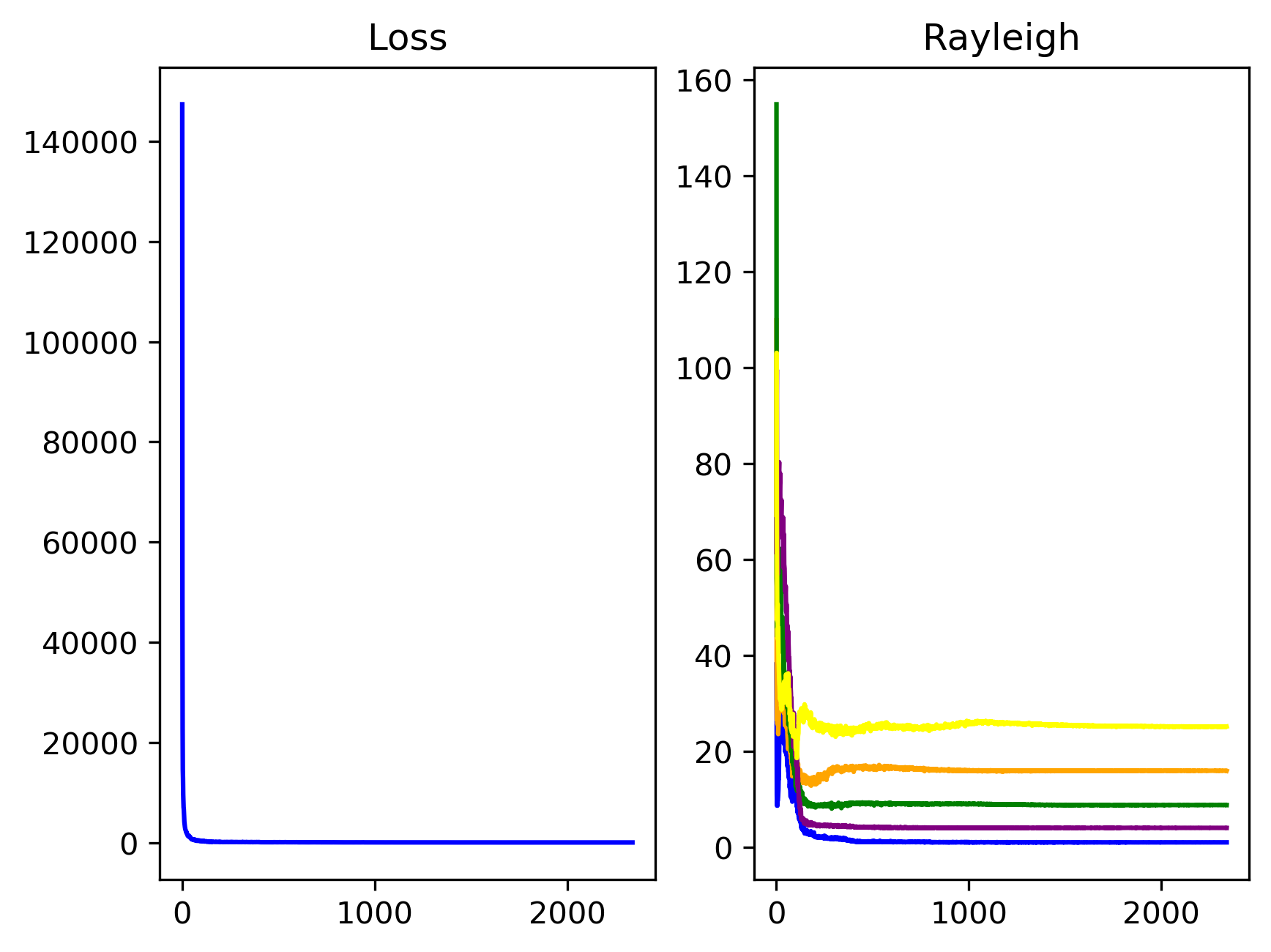}
\caption{Learning 5 different eigenpairs using factor weighted Rayleigh Quotient Penalty Loss. Three sampled epochs are shown at 50, 200, 3000 - from left to right. The network deals with the varying frequency very well, weather in the case of $\lambda=1$ where the solution is $\frac{2}{\pi}sin(x)$ to $\lambda=5$, where the solution is $\frac{2}{\pi}sin(5x)$. As before, the eigenvalues learned are, as expected: [1,4,9,16,25]. The bottom graph shows the Training Loss and the Rayleigh Quotient - mean and std($\sigma$) - per epoch.}
\end{figure}
\end{center}

\section{Implementation Details}
Our network consisted of 5 Hidden Layers, each with a varying number of neurons, from 26-50. We used $Tanh$ activation function, and an Adam optimizer with default parameters. We experimented with several known methods in Deep Learning including Dropout\cite{SrivastavaHintonEtal}, and BatchNorm\cite{IoffeSzegedy}. In our experience, we found these methods less appropriate for the given setting of learning smooth functions. We used a Starting LR(Learning Rate) of $4e-3$ and reduced it every 100 epochs by a factor of $0.7$, until the min LR of $5e-5$. We found the weight initializations\cite{8} crucial, and in our experiments a gaussian initialization with 0.0 mean and 1 std was used. The code was implemented in PyTorch and is available upon request. 

Parameter Tuning: For our final loss $\mathcal{F}_s(u)$ in eq.(\ref{eq:Fsu})
we used $\alpha=1e-1, \mu=1e-1, \delta=5e-1, \beta=1.5, c=1, R^F=1e-8, \gamma_{i}=\frac{1}{i}, \nu=2.$ Our dataset consisted of 45,000 inner-points, and 1200 boundary points. Since our boundary conditions only consists of two actual points, this can be thought of as a sort of weighing. Instead of using a true $\|\mathcal{T}u\|_{\infty}$, we used an  approximation: $\frac{\mu}{K} \sum_{k\in \text{top}_K(|{L}_i|)}^{} |{L}_k|\\ , K=40$.

\section{Summary and conclusions}
In this paper, we introduced a new method for solving the generalized eigen-problem for operators $ T: \R ^k \rightarrow \R ^n$, using Deep Neural Networks. We were able to learn an eigenfunction of an operator, given an eigenvalue $\lambda$ while keeping its norm fixed to a given value. We also showed methods of finding an eigenpair $(u, \lambda)$, given an operator $T$, using the Rayleigh Quotient. Finally, we showed a general method for finding the $k$ smallest eigenpairs $(u_{1}, \lambda_{1})...(u_{k}, \lambda_{k})$, by introducing orthogonality between each of the eigenpairs.  This is a preliminary note and much more research is needed in many directions including higher dimensions, analysis of convergence and of accuracy, comparison to other numerical methods, more complicated and non-linear operators, real life applications and more.

\end{document}